%% file: acl_latex.tex
\documentclass[11pt]{article}

\usepackage[preprint]{acl}

\usepackage{times}
\usepackage{latexsym}
\usepackage[T1]{fontenc}
\usepackage[utf8]{inputenc}
\usepackage{microtype}
\usepackage{inconsolata}

\usepackage{graphicx}
\usepackage{amsmath}
\usepackage{booktabs}
\usepackage{multirow}
\usepackage{array}
\usepackage{tabularx}
\usepackage[table]{xcolor}
\usepackage{placeins}
\usepackage{cuted}
\usepackage{needspace}
\usepackage[most]{tcolorbox}
\graphicspath{{figures/}}
\newcommand{\method}{\textsc{PersonaTree}}
\newcommand{\variant}[1]{\textit{#1}}
\definecolor{cattlerow}{HTML}{F2F2F2}
\definecolor{catline}{HTML}{C0C0C0}
\definecolor{qwenbg}{HTML}{F4F0FC}
\definecolor{geminibg}{HTML}{EEF4FF}
\definecolor{openaibg}{HTML}{F0F3F1}
\definecolor{promptbg}{HTML}{F7F8FA}
\definecolor{promptframe}{HTML}{8A96A6}
\definecolor{casebg}{HTML}{F8F7F2}
\definecolor{caseframe}{HTML}{A38B5F}
\definecolor{appendixrow}{HTML}{F6F8FA}
\definecolor{appendixhead}{HTML}{E9EEF5}
\newcommand{\modellogo}[1]{%
  \raisebox{-0.15em}{\includegraphics[height=1.05em]{#1}}%
  \hspace{0.25em}%
}

\newcommand{\best}[1]{\textbf{#1}}
\newcommand{\second}[1]{\underline{#1}}
\newcolumntype{Y}{>{\raggedright\arraybackslash}X}

\newcommand{\promptitem}[2]{\par\addvspace{0.55\baselineskip}\noindent\textbf{#1.} #2\par}
\newcommand{\casefield}[1]{\textbf{\textsc{#1}.}}
\newtcolorbox{promptbox}[1]{
  enhanced,
  breakable,
  colback=promptbg,
  colframe=promptframe,
  colbacktitle=promptframe!14,
  coltitle=black,
  fonttitle=\bfseries\small,
  fontupper=\small,
  title={#1},
  boxrule=0.45pt,
  arc=1mm,
  left=1.5mm,
  right=1.5mm,
  top=1.3mm,
  bottom=1.3mm,
  before skip=0.85\baselineskip,
  after skip=0.85\baselineskip
}
\newtcolorbox{promptpanel}[1]{
  enhanced,
  breakable,
  colback=promptbg,
  colframe=promptframe,
  colbacktitle=promptframe!16,
  coltitle=black,
  fonttitle=\bfseries\small,
  fontupper=\footnotesize\ttfamily,
  before upper={\setlength{\parskip}{0.45\baselineskip}\linespread{1.08}\selectfont\raggedright},
  title={#1},
  title after break={#1 (continued)},
  boxrule=0.55pt,
  arc=1mm,
  left=2.6mm,
  right=2.6mm,
  top=2.4mm,
  bottom=2.4mm,
  before skip=1.15\baselineskip,
  after skip=1.05\baselineskip
}
\newtcolorbox{casebox}[1]{
  enhanced,
  breakable,
  colback=casebg,
  colframe=caseframe,
  colbacktitle=caseframe!16,
  coltitle=black,
  fonttitle=\bfseries\small,
  fontupper=\small,
  before upper={\setlength{\parskip}{0.5\baselineskip}\linespread{1.1}\selectfont},
  title={#1},
  title after break={#1 (continued)},
  boxrule=0.45pt,
  arc=1mm,
  left=2.4mm,
  right=2.4mm,
  top=2mm,
  bottom=2mm,
  before skip=1.1\baselineskip,
  after skip=0.65\baselineskip
}

\title{PersonaTree: Structured Lifecycle Memory for Person Understanding in LLM Agents}

\author{
{\normalfont\small Yubo Hou\textsuperscript{1*} \quad
Jingwei Song\textsuperscript{2*} \quad
Hongbo Zhang\textsuperscript{3*} \quad
Zhisheng Chen\textsuperscript{4}} \\
{\normalfont\small Bang Xiao\textsuperscript{3} \quad
Tao Wan\textsuperscript{5} \quad
Zengchang Qin\textsuperscript{1,6\textdagger}} \\[0.2em]
{\normalfont\footnotesize \textsuperscript{1}School of ASEE, Beihang University, Beijing, China} \\
{\normalfont\footnotesize \textsuperscript{2}The University of Hong Kong, Hong Kong, China \quad
\textsuperscript{3}Peking University, Beijing, China} \\
{\normalfont\footnotesize \textsuperscript{4}University of Chinese Academy of Sciences, Beijing, China \quad
\textsuperscript{5}School of BME, Beihang University, Beijing, China} \\
{\normalfont\footnotesize \textsuperscript{6}CAIR and CECS, VinUniversity, Hanoi, Vietnam} \\
{\normalfont\footnotesize \texttt{houyubo@buaa.edu.cn} \quad \texttt{zengchang.qin@gmail.com}}
}

\begin{document}
\maketitle
\insert\footins{\footnotesize
\noindent\textsuperscript{*}Equal contribution. \quad
\textsuperscript{\textdagger}Corresponding author.\par}

\begin{abstract}
Persistent LLM agents require memory representations that make the formation of person understanding explicit across long term interaction. Existing agent memory methods emphasize information retention and retrieval, yet give limited account of how accumulated interaction evidence is abstracted into person understanding. We view this process as schema formation, where situated evidence is abstracted into reusable patterns and stable person level claims. We introduce \method, a structured lifecycle memory framework that realizes this view as a three level persona tree with explicit support paths from evidence to claims. \method{} maintains the tree through conservative writing, confidence guided consolidation, and query conditioned path retrieval, returning only the evidence depth required by each query. Across six person understanding and persistent memory benchmarks with three answer backbones, \method{} ranks first in 12 of 18 compact scores and reaches the top two in 16 settings. Ablations show that hierarchy improves abstract person understanding on KnowMe, while support path retrieval improves RealPref alignment under a comparable context budget.
\end{abstract}

\section{Introduction}
\label{sec:introduction}

LLM agents are increasingly expected to interact with the same user over extended periods of time \citep{park2023generativeagents,yao2023react,wang2023voyager,hong2024metagpt,wu2023autogen,xi2023rise}. In this setting, memory should support more than the recall of isolated facts. A useful agent needs to track what the user has said and done, recognize recurring preferences and states, and form stable judgments about the user's habits, values, constraints, and boundaries. Recent benchmarks for lifelong companions and personalized assistants reflect this shift from factual recall toward person understanding \citep{wu2026knowme,guo2026realpref,bian2026realmem,kim2025cupid}.

A central difficulty in person understanding is that abstract judgments about a user need to be grounded in scattered, situated observations. This view echoes schema theory in cognitive psychology, where schemata organize prior knowledge and relations among concepts, and guide how new observations are interpreted and recalled \citep{bartlett1932remembering,rumelhart1980schemata}. In an agent setting, this perspective motivates a memory representation that keeps the basis for user-level interpretations visible. Individual interactions may offer partial evidence, while repeated observations can gradually support more general expectations about the user's preferences, states, constraints, and boundaries. Without such grounding, an answer model may retrieve plausible user information but have limited basis for explaining why an abstract interpretation follows from the interaction history.

Existing agent memory research has made long term interaction more practical by treating memory as a problem of representation, update, organization, and retrieval. Survey work characterizes this design space across memory form, function, and lifecycle \citep{hu2026memoryage}, and recent systems instantiate it through persistent memory layers, agentic update operations, temporal organization, lightweight maintenance, and learned memory policies \citep{chhikara2025mem0,xu2025amem,zep2025graphiti,fang2025lightmem,li2026timem,li2025memos,yue2026memt}. These advances improve how user information is retained and made available during interaction. Person understanding, however, asks memory to play a stronger modeling role. The memory state should help the agent build a coherent view of the user, including preferences, states, constraints, boundaries, and motivations, rather than only supply facts, summaries, or retrieved context to the answer model. This shift calls for a representation that makes accumulated interaction experience usable for reasoning about the user as a person.

To provide such a representation, we propose \method, a structured lifecycle memory framework for person understanding in LLM agents. \method{} represents the user's evolving profile as a three level persona tree that connects interaction experience with higher level user interpretations. Leaf nodes store timestamped interaction evidence, mid level nodes capture recurring behavioral or state patterns, and root nodes represent stable persona level claims. Edges record how lower level evidence supports higher level abstractions, making the basis of each user level claim visible in the memory state. This representation turns accumulated interaction history into a structured user model that can support both concrete recall and more abstract reasoning about the person.

\method{} maintains this structured user model through lifecycle operations. Online insertion converts new interactions into typed evidence and links them to existing patterns only when schema compatibility and evidence validation support the update. Offline consolidation revisits unresolved evidence, merges compatible patterns, promotes mature abstractions into stable claims, and removes stale or weak structures. During inference, query conditioned retrieval selects the abstraction level needed for the current question and renders a compact support path when lower level evidence is useful. The answer model can therefore use the relevant claim, pattern, or event evidence without reading the full interaction history.

We evaluate \method{} on six person understanding and persistent memory benchmarks with three answer backbones. Across the full suite, \method{} ranks first in 12 of 18 compact scores and reaches the top two in 16 settings. The analysis connects these gains to the main design choices. On KnowMe, hierarchy improves questions that require reasoning from accumulated evidence to user level claims. On RealPref, retrieving a support path outperforms retrieving the same memory nodes as a flat set under a comparable context budget. Efficiency analysis further shows that \method{} keeps answer inputs compact under long user histories while improving task score.

Our contributions are as follows. First, we formulate persistent person understanding as a lifecycle memory problem in which long term interaction experience should support a coherent and grounded model of the user. Second, we introduce \method, a structured memory framework that represents user experience as a three level persona tree and maintains it through online evidence insertion, offline consolidation, confidence update, and query conditioned retrieval. Third, we provide main results, ablations, and efficiency analysis showing how hierarchical representation, support path retrieval, and context control improve persistent person understanding over extended interactions.

\section{Related Work}
\label{sec:related-work}

\paragraph{User modeling and personalization.}
Person understanding and personalization depend on long term interaction history that goes beyond isolated profile facts. Recent benchmarks make this requirement concrete through factual recall, changing states, preferences, and context dependent alignment \citep{wu2026knowme,wu2025longmemeval,guo2026realpref,kim2025cupid,bian2026realmem,li2026locomoplus,salemi2024lamp,jiang2024personallm}. MemoryBank and AI PERSONA maintain user memories or evolving profiles for personalization \citep{zhong2023memorybank,wang2024aipersona}. \method{} focuses on how such persistent user information can be organized into a coherent and grounded user model.

\paragraph{Long term agent memory.}
Long term agent memory work studies how agents store, update, filter, and retrieve memories \citep{shinn2023reflexion,sumers2024cognitive}. MemGPT is an OS-inspired virtual context manager \citep{packer2023memgpt}, while Mem0, A-MEM, LightMem, and Mem-T focus on scalable persistent memory, Zettelkasten-style linking, lightweight filtering with offline consolidation, and learned memory policies \citep{chhikara2025mem0,xu2025amem,fang2025lightmem,li2025memos,yue2026memt}. These systems are strong operational baselines, but their interfaces center on records, notes, summaries, graphs, or retrieved contexts. \method{} organizes memory as a structured user model that preserves the basis for higher level interpretation.

\paragraph{Structured and hierarchical memory.}
Structured memory methods go beyond flat records by organizing long histories with relations or hierarchy \citep{lewis2021rag,liu2023lostmiddle,sarthi2024raptor,edge2025graphrag,gutierrez2025hipporag}. Zep/Graphiti uses temporal knowledge graphs \citep{zep2025graphiti}, TiMem builds a temporal memory tree \citep{li2026timem}, and MemTree builds dynamic tree schemas \citep{rezazadeh2024memtree}. These structures support entity, temporal, topical, or schema organization. \method{} uses hierarchy for evidential abstraction in person understanding, linking interaction events, recurring patterns, and persona claims so abstract user interpretations remain grounded.

\section{Method}
\label{sec:method}

\subsection{Overview}
\label{sec:method-overview}

\method{} represents long term user memory as a three level persona tree. Leaves store timestamped interaction evidence, mid nodes capture recurring behavior or state patterns, and roots store stable persona claims. Typed support edges keep higher level claims traceable to the evidence below them, allowing the same memory state to support factual recall and abstract person understanding.

\begin{figure*}[t]
  \centering
  \includegraphics[width=.98\textwidth]{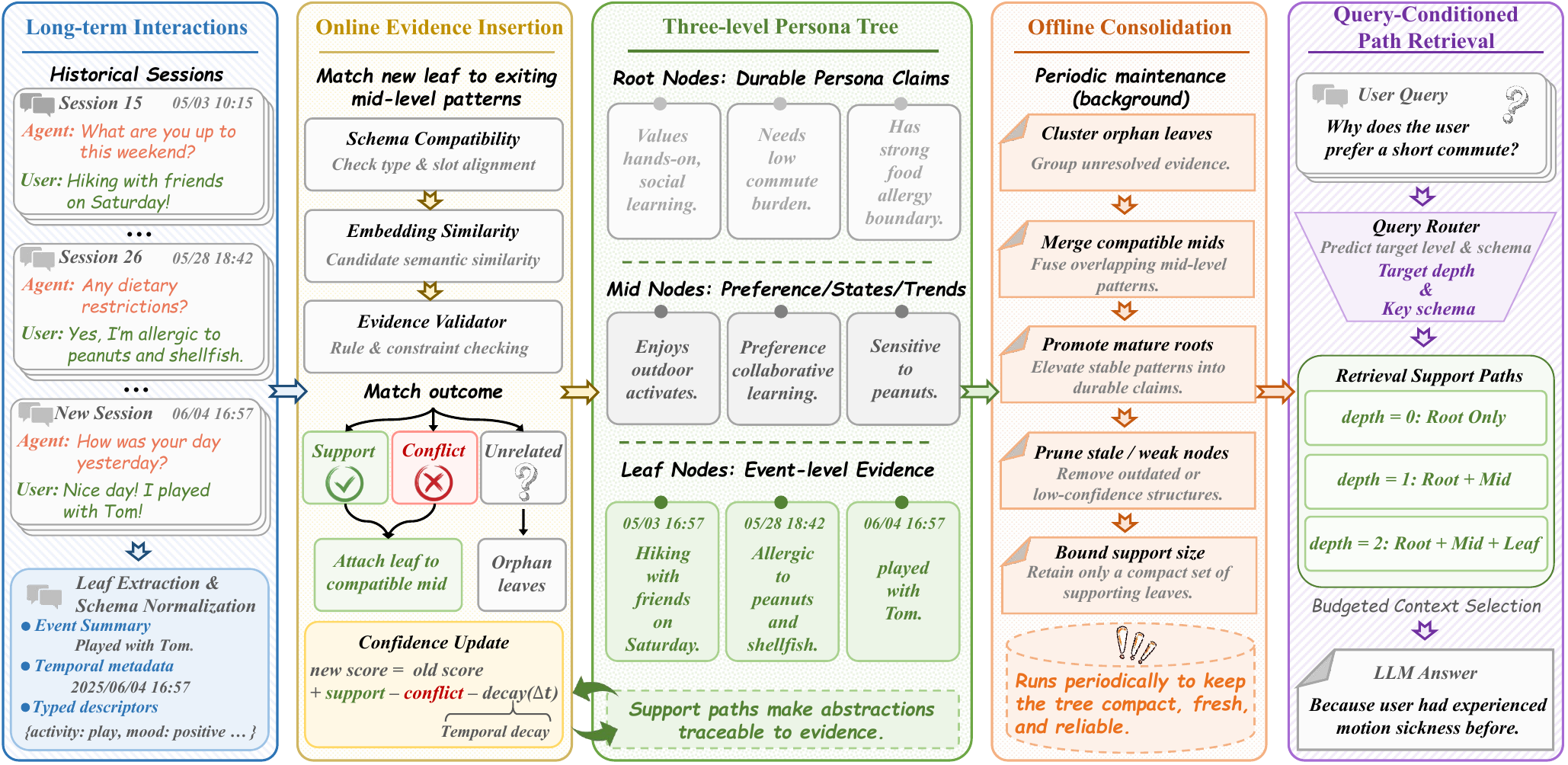}
  \caption{Overview of \method{}. The framework converts long term interactions into evidence leaves, consolidates reusable patterns into mid level nodes, promotes stable user claims into roots, and retrieves compact support paths for query conditioned answering.}
  \label{fig:method-overview}
\end{figure*}

\method{} updates this tree through a lifecycle of writing, consolidation, and retrieval. New interactions first enter as typed leaves and attach to existing abstractions only when schema compatibility and evidence validation support the link. Later consolidation merges compatible evidence, promotes mature abstractions, and removes stale structure. At inference time, retrieval chooses the abstraction level required by the query and renders only the needed support depth. Figure~\ref{fig:method-overview} summarizes the full lifecycle.

\subsection{Lifecycle Memory State}
\label{sec:lifecycle-memory-state}

\method{} defines memory as a typed graph whose support edges form a three level persona tree. A \textbf{Leaf} node records event level evidence from one interaction or a short local segment. Above the leaves, \textbf{Mid} nodes capture reusable behavior or state abstractions, and \textbf{Root} nodes represent durable user claims grounded in those abstractions. The support edges connect lower level evidence to higher level interpretation, so each abstract claim remains traceable to the observations that justify it.

Formally, for user \(u\), we write the memory state as
\begin{equation}
  \mathcal{T}_u=(\mathcal{V}^{L}_u,\mathcal{V}^{M}_u,\mathcal{V}^{R}_u,\mathcal{E}_u),
  \label{eq:personatree-state}
\end{equation}
where \(\mathcal{V}^{L}_u\), \(\mathcal{V}^{M}_u\), and \(\mathcal{V}^{R}_u\) denote leaf, mid, and root nodes, and \(\mathcal{E}_u\) contains typed support edges. Each node \(v=(x,t,a,z,c,\ell)\) stores textual content, temporal metadata, task attributes, an embedding, confidence, and an abstraction level \(\ell\in\{L,M,R\}\).

The attribute vector \(a\) stores schema information for the target application, including domains, event types, states, roles, and boundaries. In our experiments, these attributes instantiate persona oriented descriptors for person understanding. The lifecycle operators remain defined over abstraction levels and support edges, allowing the same memory state to be instantiated with other schemas.

\subsection{Online Evidence Insertion}
\label{sec:online-writing}

Online insertion adds new interactions to the tree as evidence before they become abstractions. Given a new interaction \(h_t\), \method{} extracts a typed leaf \(l_t\) with textual content, temporal metadata, schema attributes, and an embedding. Before insertion, normalization under the target schema maps factual updates, subjective preferences, and safety constraints into compatible attribute types.

The new leaf is then matched against existing mid nodes. Candidate mids first need to be schema compatible with the leaf. Among compatible candidates, \method{} combines embedding similarity with an evidence validator, since semantic closeness alone does not determine whether a leaf supports an abstraction. The validator labels the relation as support, conflict, or unrelated, and these judgments are mapped to positive, negative, or near-zero evidence scores. The match score is
\begin{equation}
\begin{aligned}
  s_{\mathrm{match}}(l,m)
  &= \mathbf{1}[\kappa(a_l,a_m)=1]\, s(l,m), \\
  s(l,m)
  &= \alpha \cos(z_l,z_m)+\beta\,\psi(l,m),
\end{aligned}
\label{eq:match-score}
\end{equation}
where \(\kappa(a_l,a_m)\) is a compatibility predicate under the schema and \(\psi(l,m)\in[-1,1]\) maps the validator judgment from conflict to support. If \(\max_m s_{\mathrm{match}}(l,m)\ge \theta_M\), the leaf attaches to the best mid. When the validator finds a conflict, the leaf remains separate from the candidate abstraction and passes its negative evidence weight to the confidence update. If no candidate passes the threshold, the leaf remains orphaned until consolidation provides enough context for a safer abstraction.

\subsection{Confidence Update and Short-Term Activation}
\label{sec:confidence-activation}

\method{} uses confidence as a shared signal that carries supporting and conflicting evidence into later selection, pruning, and promotion decisions. For a matched node, confidence is updated in log-odds space with temporal decay and weights derived from evidence:
\begin{align}
  L_t &=
  (L_{t-1} - L_{\mathrm{base}}) e^{-\lambda \Delta t}
  + L_{\mathrm{base}} + \omega_E, \nonumber \\
  c_t &= \sigma(L_t),
  \label{eq:confidence-update}
\end{align}
where \(L_t\) is the current log-odds, \(L_{\mathrm{base}}\) is a baseline prior, \(\Delta t\) is the elapsed time since the last update, \(\lambda\) is the decay rate for the node type, and \(\omega_E\) is the evidence weight assigned by the validator.

The log-odds form lets support and conflict act as positive and negative evidence weights, while the sigmoid mapping keeps the resulting confidence bounded. The decay term pulls confidence toward the baseline when no new evidence arrives, while \(\omega_E\) carries support or contradiction from the validator. Schema types whose content should persist can use reduced or zero decay; less persistent behavioral patterns fade without supporting evidence. \method{} also keeps recent activations for updated or retrieved nodes, giving retrieval a recency signal separate from longer term reliability.

\subsection{Offline Consolidation}
\label{sec:offline-consolidation}

Offline consolidation handles evidence that is too sparse or unstable for online abstraction. It periodically revisits unresolved leaves and low confidence abstractions. First, it forms candidate clusters from orphan leaves using schema attributes, temporal proximity, and semantic similarity. A cluster becomes a mid candidate only when it contains enough evidence and its leaves can support a consistent pattern. If the pattern expressed by the cluster matches an existing mid, \method{} attaches the member leaves to that mid and rewrites its description to reflect the combined evidence. Otherwise, \method{} abstracts the cluster into a new mid node and links the member leaves as its support.

Root promotion applies stricter thresholds than consolidation at the mid level. \method{} first selects mids that pass thresholds for confidence, support size, and temporal coverage. It then groups compatible mids into candidate root clusters and summarizes each cluster into a durable user claim. These promotion criteria make root formation conservative, so weak or narrowly contextual evidence is not elevated into durable claims about the user.

The same pass also limits memory growth. Stale nodes decay, old orphan leaves are removed, and mids retain only a bounded set of supporting leaves. The system prunes mids and roots with low confidence. When a weak root is deleted, its mids are detached, while confident local patterns remain available.

\subsection{Query-Conditioned Path Retrieval}
\label{sec:intent-retrieval}

Retrieval is conditioned on the granularity of the query. A query may ask for a concrete event, a recurring pattern or evolving state, or a durable claim about the user. \method{} therefore routes the query \(q\) to an abstraction level \(h\in\{L,M,R\}\) and predicts schema attributes that constrain the search space:
\begin{equation}
\begin{aligned}
  \hat{h} &= \arg\max_{h\in\{L,M,R\}} p_\phi(h\mid q), \\
  \hat{a}_q &= g_\phi(q),
\end{aligned}
\label{eq:intent-routing}
\end{equation}
where \(p_\phi\) is the routing model, \(\hat{h}\) is the selected level, and \(g_\phi\) predicts the query attribute constraint \(\hat{a}_q\).

After routing, \method{} renders only the evidence depth needed by the query and the token budget. A leaf level query is answered from compact event memories. For mid and root level queries, retrieval starts with the shallowest useful rendering and expands along the support path when the answer needs event detail, temporal evidence, or an explicit rationale. Given retrieved support paths \(\mathcal{P}_q\), where each path \(P\) connects a root, a supporting mid, and leaves when available, the renderer \(g_d(P)\) exposes \(P\) as a root alone (\(d=0\)), a root with its supporting mid (\(d=1\)), or the full root--mid--leaf chain (\(d=2\)). The final context is selected under a token budget:
\begin{align}
  \mathcal{G}_q
  &= \{g_d(P): P\in\mathcal{P}_q,\ d\in\{0,1,2\}\}, \nonumber\\
  \mathcal{C}^{\star}
  &= \arg\max_{\mathcal{C}\subseteq \mathcal{G}_q}
    \sum_{x\in\mathcal{C}}
    \big(r(q,x)-\lambda_b \mathrm{tok}(x)\big), \nonumber\\
  \text{s.t.}\quad
  &\sum_{x\in\mathcal{C}}\mathrm{tok}(x)\le B_q,
  \label{eq:budgeted-disclosure}
\end{align}
where \(\mathcal{G}_q\) is the set of candidate renderings, \(\mathcal{C}^{\star}\) is the selected context, \(B_q\) is the token budget, \(\mathrm{tok}(x)\) counts the tokens in rendering \(x\), and \(\lambda_b\) weights token cost. The score \(r(q,x)\) favors a rendering when it is aligned with the query, supported by reliable or recently activated nodes, and set at the evidence depth requested by the query. The token penalty keeps the context compact unless deeper evidence is needed to answer or justify the query.

\subsection{Person Understanding Instantiation}
\label{sec:persona-instantiation}

For person understanding, we instantiate the generic attribute vector with persona oriented descriptors. Leaf nodes store local evidence fields, including interaction domain, temporal metadata, typed descriptors, and a compact event summary. Mid nodes organize compatible leaves into reusable person patterns such as preferences, ongoing states, recurring triggers, and temporal trends. Root nodes summarize durable user claims over identity, personality, values, principles, and safety relevant boundaries. This schema defines the experimental instantiation of \method{}, while the lifecycle state, confidence update, consolidation, and path retrieval mechanisms remain unchanged under alternative schemas. Appendix~\ref{sec:appendix-schema} lists the concrete schema and implementation rules.

\section{Experiments}
\label{sec:experiments}

We evaluate \method{} across six person understanding and persistent memory benchmarks, testing whether its lifecycle improves user modeling while remaining practical under long interaction histories.

\begin{table*}[t]
\centering
\setlength{\aboverulesep}{0pt}
\setlength{\belowrulesep}{0pt}
\setlength{\tabcolsep}{5.4pt}
\renewcommand{\arraystretch}{1.05}
\caption{\textbf{Main results across person understanding and persistent memory benchmarks.} Scores are compact summaries computed from the native metrics of each benchmark; higher is better. Strong reference rows use \(K=V+\mathrm{fact}\) on LongMemEval, Graphiti on RealMem, and SeCom on LoCoMo-Plus. Appendix~\ref{sec:appendix-evaluation-details} defines each aggregation. Best results are \best{bolded}, and second best results are \second{underlined}.}
\label{tab:main-results}
\small
\begin{tabular}{@{}lcccccc@{}}
\toprule
\multicolumn{7}{@{}>{\columncolor{qwenbg}[0pt][0pt]}l@{}}{\modellogo{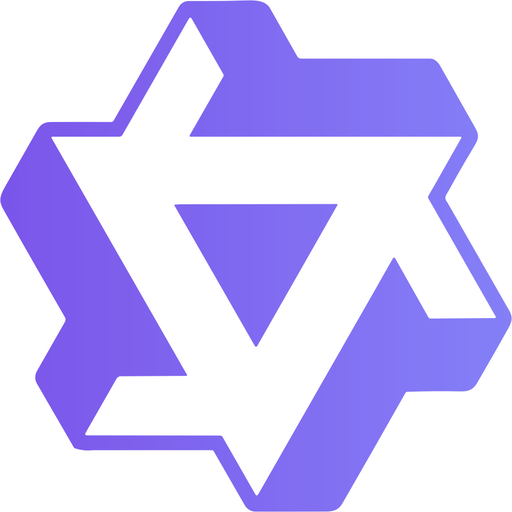}\textbf{Qwen3-32B}} \\
\midrule
\textbf{Method} & \textbf{KnowMe} & \textbf{LongMemEval} & \textbf{RealPref} & \textbf{RealMem} & \textbf{CUPID} & \textbf{LoCoMo-Plus} \\
\midrule
Full history & 39.9 & -- & 60.1 & 39.1 & 48.8 & 20.1 \\
Flat retrieval & 43.3 & 68.1 & 73.5 & 40.3 & 50.1 & 20.4 \\
Benchmark strong & -- & 72.1 & -- & \second{46.2} & -- & 15.3 \\
Mem0 & 43.2 & 68.6 & 75.6 & 44.4 & 51.8 & 16.1 \\
A-MEM & 44.8 & 70.0 & 75.8 & 40.5 & 51.7 & 17.7 \\
TiMem & \second{46.1} & \best{72.7} & \second{76.9} & 44.3 & \second{53.1} & \second{20.9} \\
\method{} & \best{47.6} & \second{72.3} & \best{78.1} & \best{47.8} & \best{55.8} & \best{22.3} \\
\midrule
\multicolumn{7}{@{}>{\columncolor{geminibg}[0pt][0pt]}l@{}}{\modellogo{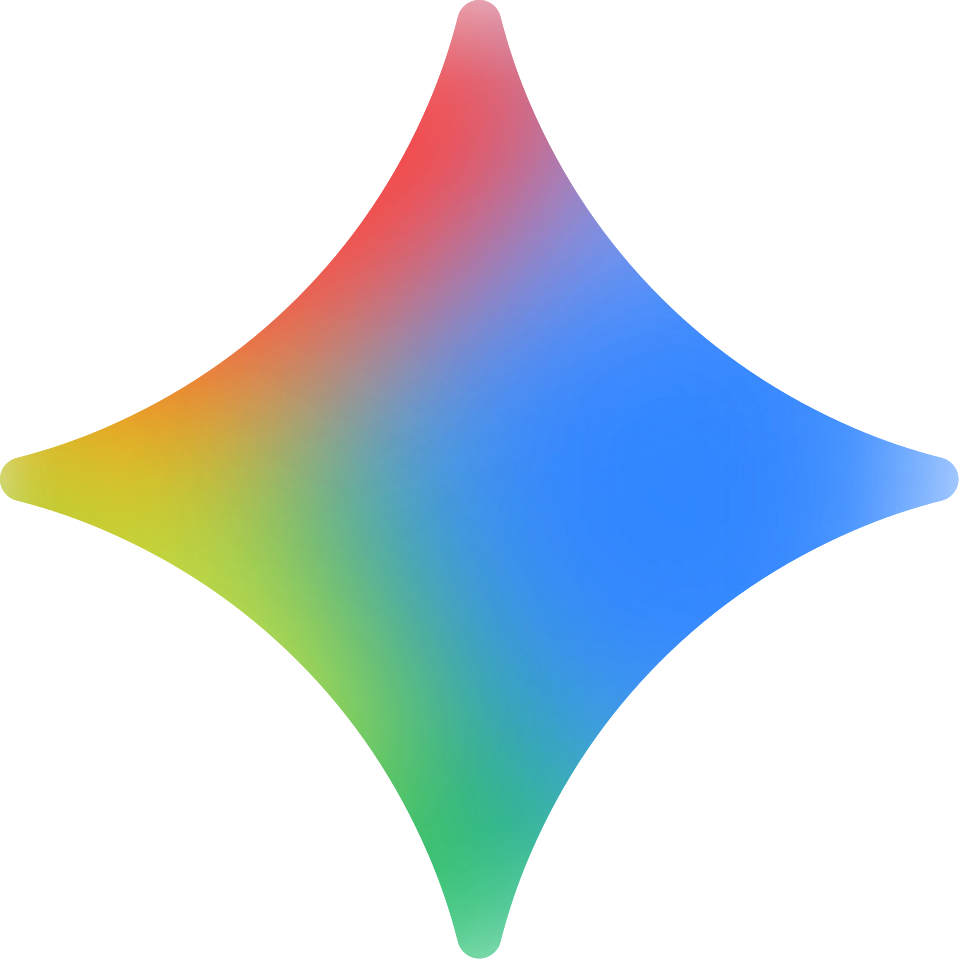}\textbf{Gemini 3 Flash}} \\
\midrule
\textbf{Method} & \textbf{KnowMe} & \textbf{LongMemEval} & \textbf{RealPref} & \textbf{RealMem} & \textbf{CUPID} & \textbf{LoCoMo-Plus} \\
\midrule
Full history & 45.4 & -- & 63.5 & 40.4 & 55.1 & \second{24.8} \\
Flat retrieval & 48.3 & 69.1 & 80.6 & 41.5 & 56.7 & \best{25.2} \\
Benchmark strong & -- & 73.4 & -- & \best{47.4} & -- & 17.8 \\
Mem0 & 47.8 & 70.0 & 82.7 & 45.6 & \second{59.5} & 18.3 \\
A-MEM & 49.4 & 70.9 & 82.5 & 41.6 & 57.4 & 20.3 \\
TiMem & \best{50.7} & \second{73.6} & \best{83.1} & 45.1 & 59.1 & 23.6 \\
\method{} & \second{50.4} & \best{76.2} & \second{83.0} & \second{47.1} & \best{61.4} & 24.6 \\
\midrule
\multicolumn{7}{@{}>{\columncolor{openaibg}[0pt][0pt]}l@{}}{\modellogo{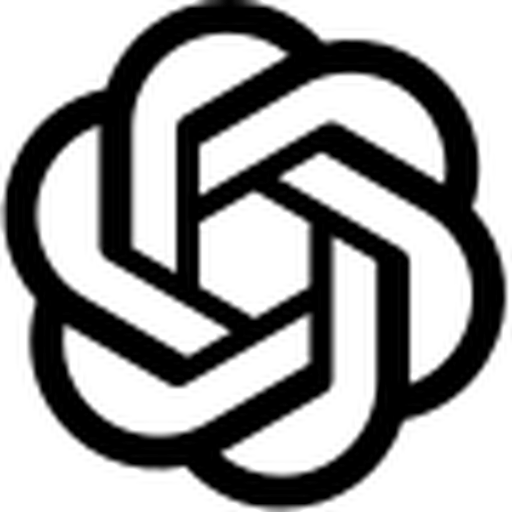}\textbf{GPT-5.4 Mini}} \\
\midrule
\textbf{Method} & \textbf{KnowMe} & \textbf{LongMemEval} & \textbf{RealPref} & \textbf{RealMem} & \textbf{CUPID} & \textbf{LoCoMo-Plus} \\
\midrule
Full history & 46.8 & -- & 77.5 & 41.3 & 56.8 & 22.9 \\
Flat retrieval & 49.9 & 70.3 & 85.7 & 42.4 & 58.3 & 23.6 \\
Benchmark strong & -- & \second{74.5} & -- & \second{48.2} & -- & 18.2 \\
Mem0 & 49.1 & 71.7 & 87.0 & 46.6 & 59.5 & 18.8 \\
A-MEM & 50.9 & 72.3 & \second{88.1} & 42.9 & \second{60.8} & 20.7 \\
TiMem & \second{52.1} & \best{74.9} & 87.9 & 46.1 & 60.6 & \second{24.2} \\
\method{} & \best{53.3} & 74.3 & \best{89.1} & \best{49.8} & \best{63.5} & \best{25.5} \\
\bottomrule
\end{tabular}
\end{table*}

\subsection{Experimental Setup}
\label{sec:experimental-setup}

The suite combines person understanding and preference alignment benchmarks (KnowMe, RealPref, CUPID) with long term memory benchmarks for chat histories, project oriented interaction, and beyond factual cognitive memory (LongMemEval, RealMem, LoCoMo-Plus) \citep{wu2026knowme,guo2026realpref,kim2025cupid,wu2025longmemeval,bian2026realmem,li2026locomoplus}. Baselines cover full history prompting, flat retrieval, persistent memory systems, and strong references from individual benchmarks when available. Table~\ref{tab:main-results} reports one compact summary score for each benchmark with Qwen3-32B, Gemini 3 Flash, and GPT-5.4 Mini as answer backbones. Appendix~\ref{sec:appendix-evaluation} describes the benchmark scope, baseline policy, compact score definitions, and metric breakdown.

\subsection{Main Results}
\label{sec:main-results}

Across six benchmarks and three answer backbones, \method{} gives the strongest overall pattern in Table~\ref{tab:main-results}, ranking first in 12 of 18 compact scores and reaching the top two in 16 settings. Its advantage is most consistent on benchmarks that require stable user modeling and preference use. On CUPID, \method{} is best under all three answer backbones, ahead of the next strongest row by 2.7, 1.9, and 2.7 points. On RealPref and KnowMe, it leads with Qwen3-32B and GPT-5.4 Mini and remains top two with Gemini 3 Flash. Taken together, these results suggest that the lifecycle representation helps models accumulate user evidence and reuse it across preference and persona questions.

The broader memory benchmarks show a similar pattern against persistent memory systems and benchmark strong references. On RealMem and LoCoMo-Plus, \method{} leads with Qwen3-32B and GPT-5.4 Mini. On LongMemEval, it obtains the highest Gemini 3 Flash score and remains top two with Qwen3-32B. Across answer models, Qwen3-32B and GPT-5.4 Mini each place \method{} first on five of six benchmarks, while Gemini 3 Flash gives two leading scores and three additional top two scores.

\subsection{Efficiency under Long User Histories}
\label{sec:efficiency-analysis}

\begin{center}
  \centering
  \includegraphics[width=.95\columnwidth]{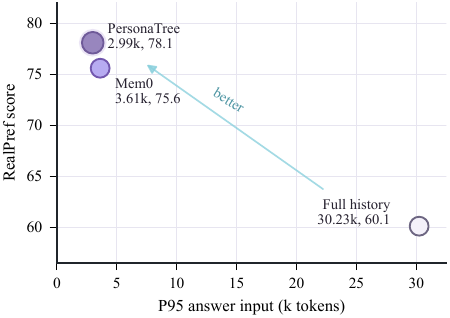}
  \captionof{figure}{RealPref score and context size. \method{} attains a higher task score than full history prompting and Mem0 while keeping the P95 answer input below 3k tokens.}
  \label{fig:realpref-efficiency-tradeoff}
\end{center}

\begin{figure*}[!t]
  \centering
  \captionsetup{skip=2pt}
  \includegraphics[width=.92\textwidth]{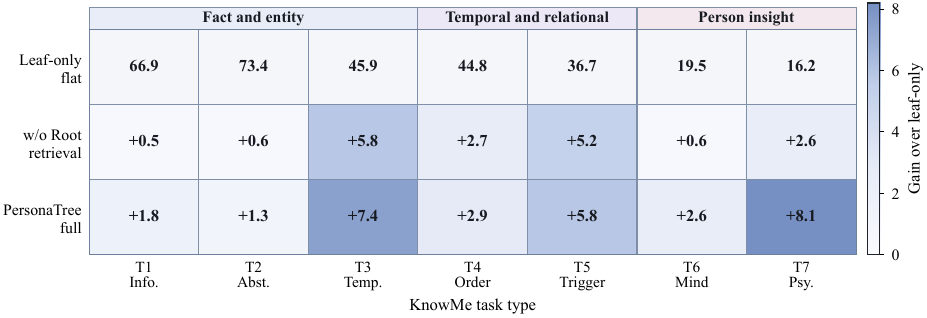}
  \caption{KnowMe hierarchy ablation across task types. The leaf only row reports the original score, while the other rows report gains over retrieval from leaves alone.}
  \label{fig:knowme-hierarchy-gain}
\end{figure*}

Long user histories make answer context size part of the evaluation. A memory method should improve preference following without sending an expanding interaction history to the answer model. Table~\ref{tab:realpref-efficiency-main} reports the RealPref score together with median retrieved tokens, P95 answer input, input growth per 100 turns, and the gap between median and P95 input. The final columns record median synchronous latency and background update tokens.

\begin{table}[t]
\centering
\setlength{\aboverulesep}{0pt}
\setlength{\belowrulesep}{0pt}
\setlength{\tabcolsep}{5.2pt}
\renewcommand{\arraystretch}{1.16}
\small
\caption{\textbf{RealPref efficiency statistics.} Token counts are shown in thousands. Latency reports median added latency.}
\label{tab:realpref-efficiency-main}
\begin{tabular}{>{\bfseries\raggedright\arraybackslash}lccc}
\toprule
\rowcolor{appendixhead}
\textbf{Metric} & \textbf{Full history} & \textbf{Mem0} & \textbf{\method{}} \\
\midrule
Score & 60.1 & 75.6 & \best{78.1} \\
Ret. P50 & 0.00k & 1.71k & \best{1.34k} \\
Input P95 & 30.23k & 3.61k & \best{2.99k} \\
Growth & 24.46k & 0.69k & \best{0.27k} \\
Gap & 6.09k & 1.67k & \best{1.42k} \\
Latency & 0.00s & 0.18s & 0.35s \\
Bg./Turn & 0.00k & 0.81k & 0.94k \\
\bottomrule
\end{tabular}
\end{table}

Compared with full history prompting, \method{} improves the RealPref score from 60.1 to 78.1 while reducing the P95 answer input from 30.23k to 2.99k tokens and the input growth from 24.46k to 0.27k tokens per 100 turns. Compared with Mem0, it retrieves fewer median tokens, lowers the P95 answer input, and improves the score from 75.6 to 78.1. The input gap also drops to 1.42k tokens, indicating a more stable answer context across typical and high load turns. Figure~\ref{fig:realpref-efficiency-tradeoff} plots score against answer context length.

\section{Ablation and Mechanism Analysis}
\label{sec:analysis}

\begin{figure*}[!t]
  \centering
  \captionsetup{skip=2pt}
  \includegraphics[width=.92\textwidth]{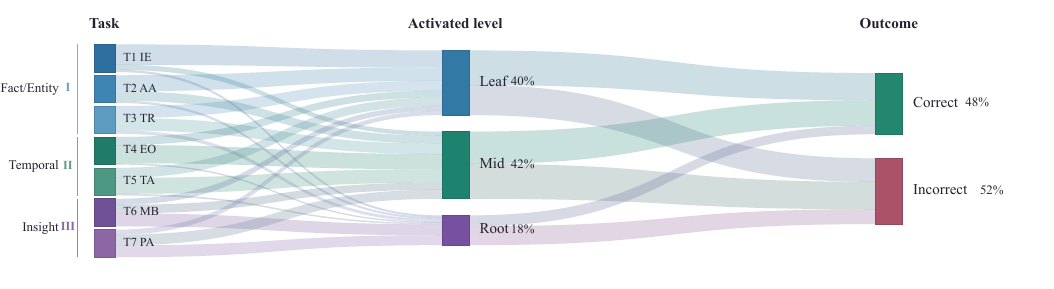}
  \caption{KnowMe flow from task type to activated memory level and answer outcome.}
  \label{fig:activation-sankey}
\end{figure*}

The main results establish the outcome pattern; we now examine which memory behaviors help explain it. All analyses use Qwen3-32B and keep the data, judge, decoding setup, and answer format fixed, so the comparisons isolate changes in memory organization and retrieval policy. The analyses focus on how hierarchy, support paths, and retrieval depth help the answer model connect local observations with higher level interpretations about the user. Appendix~\ref{sec:appendix-full-analysis-results} reports the full ablation and activation tables behind the compact results discussed here.

\begin{table}[t]
\centering
\setlength{\aboverulesep}{0pt}
\setlength{\belowrulesep}{0pt}
\setlength{\tabcolsep}{2.2pt}
\renewcommand{\arraystretch}{1.12}
\footnotesize
\caption{\textbf{Core ablations.} Compact summary of hierarchy and path-retrieval ablations under Qwen3-32B.}
\label{tab:core-ablation-main}
\newcommand{\keystack}[2]{\begin{tabular}[c]{@{}c@{}}#1\\[0.3ex]#2\end{tabular}}
\newcommand{\settingcell}[1]{\scriptsize\bfseries #1}
\makebox[\columnwidth][c]{\resizebox{0.985\columnwidth}{!}{%
\begin{tabular}{>{\centering\arraybackslash}m{0.15\columnwidth}
                >{\bfseries\raggedright\arraybackslash}m{0.27\columnwidth}
                >{\centering\arraybackslash}m{0.10\columnwidth}
                >{\centering\arraybackslash}m{0.22\columnwidth}
                >{\centering\arraybackslash}m{0.10\columnwidth}
                >{\centering\arraybackslash}m{0.10\columnwidth}}
\toprule
\rowcolor{appendixhead}
\textbf{Setting} & \textbf{Variant} & \textbf{Score} & \textbf{Key scores} & \textbf{Ctx.} & \textbf{Gain} \\
\midrule
\multirow{3}{*}[-4.1ex]{\settingcell{KnowMe}}
& Leaf Only & 43.3 & \keystack{T6 19.5}{T7 16.2} & -- & +0.0 \\
\cmidrule{2-6}
& No Root & 45.9 & \keystack{T6 20.1}{T7 18.8} & -- & +2.6 \\
\cmidrule{2-6}
& \method{} & \best{47.6} & \keystack{\best{T6 22.1}}{\best{T7 24.3}} & -- & \best{+4.3} \\
\midrule
\multirow{2}{*}[-2.4ex]{\settingcell{RealPref}}
& Flat Nodes & 75.7 & \keystack{Align. 3.61}{Qual. 4.00} & 3.18k & +0.0 \\
\cmidrule{2-6}
& \method{} Path & \best{78.1} & \keystack{\best{Align. 3.91}}{\best{Qual. 4.20}} & \best{2.99k} & \best{+2.4} \\
\bottomrule
\end{tabular}}}
\end{table}

\subsection{Hierarchy Ablation on Abstract Person Understanding}
\label{sec:hierarchy-ablation}

KnowMe lets us examine hierarchy across different depths of person understanding, since its tasks range from local factual questions to abstract person insight. Under the same answer backbone and budget, \variant{Leaf Only} retrieves raw leaf evidence, \variant{No Root Retrieval} keeps refinement from leaves to mid nodes while root retrieval is disabled, and \method{} retrieves over Leaf, Mid, and Root paths.

Abstract person questions require the model to synthesize a higher level interpretation from dispersed observations, and the hierarchy mirrors this inferential structure. Leaf evidence contains the observations, mid nodes collect repeated behavior and evolving states, and root nodes store durable persona hypotheses backed by those patterns. Mid nodes reduce the risk that a root claim is inferred directly from isolated events, while root nodes provide reusable interpretations that can be applied across questions. This gives the answer model a scaffold for reasoning about stable preferences, boundaries, and latent intent while retaining the lower level support needed for grounded answers.

The full hierarchy raises the KnowMe macro average from 43.3 to 47.6, a \(+4.3\)-point gain over retrieval from leaves alone. The largest change appears on T7, from 16.2 to 24.3, where questions written by expert annotators require reasoning about complex motivations and identity construction from dispersed narrative evidence. Figure~\ref{fig:knowme-hierarchy-gain} shows that the gains also extend to T3--T5, which KnowMe defines as temporal reasoning, logical event ordering, and mnestic trigger analysis. Table~\ref{tab:core-ablation-main} reports the compact scores behind these changes.

\subsection{Path Retrieval Ablation on Preference Following}
\label{sec:path-ablation}

After isolating the role of hierarchy, we next ask whether the retrieved structure itself matters when the available memory nodes are held fixed. The flat variant ranks all Leaf, Mid, and Root nodes in one pool, while \method{} returns support chains that connect an abstract preference to the observations that formed it.

\begin{center}
  \centering
  \includegraphics[width=.97\columnwidth]{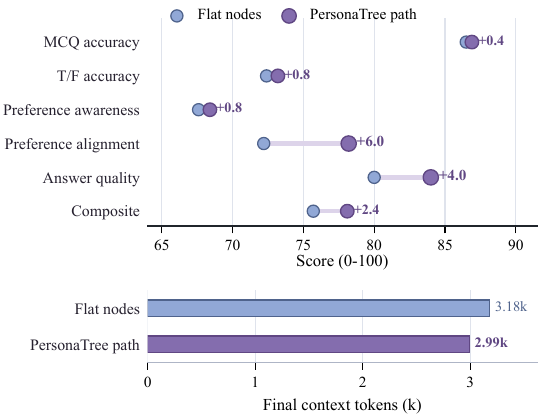}
  \captionof{figure}{RealPref path retrieval ablation. Support path retrieval yields stronger preference alignment and answer quality with a final context of 2.99k tokens.}
  \label{fig:realpref-path-ablation}
\end{center}

Preference following requires a grounded representation of both the preference and its support. A response often depends on what the system believes about the user, why that belief is plausible, and which observations support it in the current situation. Path retrieval keeps these pieces together in the returned context. In the chain, observed behavior supports a recurring preference, and the preference then constrains the answer.

Path retrieval improves the RealPref composite score from 75.7 to 78.1, as summarized in Table~\ref{tab:core-ablation-main} and Figure~\ref{fig:realpref-path-ablation}. Most of the gain comes from generation metrics. Preference alignment rises from 3.61 to 3.91, and answer quality rises from 4.00 to 4.20. Because the final contexts are kept at similar sizes, 3.18k tokens for flat retrieval over all nodes and 2.99k tokens for support path retrieval, the result suggests that the improvement comes from how evidence is organized in the retrieved context.

\subsection{Memory Activation by Task Type}
\label{sec:layer-activation}

The activation analysis examines how \method{} moves through the memory hierarchy when answering different kinds of person understanding questions. As the question shifts from factual recall to temporal organization and then to person insight, the system should rely on increasingly abstract memory levels while keeping lower level evidence available for inspection. We therefore use activation traces to study whether the retrieved levels follow this change in question demand.

Figure~\ref{fig:activation-sankey} connects task types, activated memory levels, and answer outcomes. T1/T2 mostly activate Leaf nodes, T4/T5 shift toward Mid nodes, and T6/T7 activate Root nodes much more often. This distribution is consistent with the intended division of labor in the hierarchy. Local factual questions rely on event evidence, temporal and narrative questions use mid level organization, and insight questions more often require root level abstractions. The retrieved support paths then expose the lower level evidence behind these abstractions, making the resulting user interpretation easier to inspect.

\section{Conclusion}
\label{sec:conclusion}

This paper framed persistent person understanding as a lifecycle memory problem and introduced \method{}, a three level persona tree that connects event evidence, recurring patterns, and durable user claims through typed support edges. By combining conservative writing, evidence based consolidation, confidence update, and path retrieval, \method{} keeps abstract user interpretations grounded in the observations that support them.

Across six person understanding and persistent memory benchmarks, \method{} ranks first in 12 of 18 compact scores and reaches the top two in 16 settings. Ablations show that hierarchy supports abstraction from dispersed evidence, while support path retrieval improves preference following under a comparable context budget. Together, these results support a view of agent memory as a structured evidential basis for forming, retrieving, and inspecting user models over time.

\section*{Limitations}

This work focuses on text based interaction histories and evaluates \method{} on English language benchmarks. Extending the same lifecycle representation to multilingual settings, speech interaction, and multimodal companion scenarios would test how the schema should adapt when user evidence arrives through richer channels.

\section*{Ethics Statement}

This paper evaluates \method{} on existing research benchmarks and does not involve deployment with real users. For practical use, persistent memory systems should follow standard data governance practices, including transparency about what is stored, user control over retained information, and appropriate safeguards for personal data.

\nocite{park2023generativeagents,yao2023react,wang2023voyager,hong2024metagpt,wu2023autogen,xi2023rise,li2025memos,salemi2024lamp,jiang2024personallm,shinn2023reflexion,sumers2024cognitive,guu2020realm,karpukhin2020dpr,lewis2021rag,izacard2021fid,borgeaud2022retro,izacard2022atlas,gao2022hyde,jiang2023flare,liu2023lostmiddle,bai2024longbench,sarthi2024raptor,edge2025graphrag,gutierrez2025hipporag,asai2023selfrag,maharana2024locomo,tan2025membench,hu2026memoryagentbench}
\bibliography{custom}

\clearpage
\appendix
\setlength{\parskip}{0.28\baselineskip}

\section{Evaluation Details and Benchmark Results}
\label{sec:appendix-evaluation}

\subsection{Evaluation Details}
\label{sec:appendix-evaluation-details}

\paragraph{Benchmarks.}
The benchmark suite combines person understanding, preference alignment, and long term memory evaluation. KnowMe measures person understanding across seven task types, from factual questions to high level person insight \citep{wu2026knowme}. RealPref measures whether an assistant can infer and follow long horizon user preferences \citep{guo2026realpref}, while CUPID tests contextualized preference alignment from interactions \citep{kim2025cupid}. LongMemEval evaluates long term interactive memory with retrieval and QA metrics \citep{wu2025longmemeval}. RealMem evaluates memory driven interaction in realistic project settings \citep{bian2026realmem}. LoCoMo-Plus evaluates beyond factual cognitive memory in long dialogues \citep{li2026locomoplus}.

\paragraph{Baselines.}
The comparisons separate raw history prompting, unstructured retrieval, and persistent memory systems. Full history prompting measures how far the answer model can use the accumulated interaction directly. Flat retrieval measures the effect of selecting relevant snippets without the hierarchy and support links used by \method{}. The main persistent memory baselines are Mem0 \citep{chhikara2025mem0}, A-MEM \citep{xu2025amem}, and TiMem \citep{li2026timem}. We include specialized references when a benchmark provides a directly comparable setting: \(K=V\) and \(K=V+\mathrm{fact}\) for LongMemEval, Graphiti/Zep for RealMem \citep{zep2025graphiti}, and SeCom for LoCoMo-Plus.

\paragraph{Metrics.}
The main table uses one compact score per benchmark for readability. Table~\ref{tab:compact-score-definitions} defines the aggregation. The compact scores are computed from the detailed metric columns; accuracy, F1, retrieval, and QA metrics are already on a 0--100 scale, while 1--5 judge scores are scaled to the same range before averaging.

\begin{table*}[t]
\centering
\setlength{\aboverulesep}{0pt}
\setlength{\belowrulesep}{0pt}
\setlength{\tabcolsep}{5.5pt}
\renewcommand{\arraystretch}{1.2}
\small
\caption{\textbf{Compact score aggregation.} Higher values are better after the listed scaling.}
\label{tab:compact-score-definitions}
\begin{tabularx}{\textwidth}{>{\bfseries\raggedright\arraybackslash}p{0.13\textwidth}YY}
\toprule
\rowcolor{appendixhead}
\textbf{Benchmark} & \textbf{Metric columns} & \textbf{Compact score} \\
\midrule
KnowMe & \(T_1\)--\(T_7\) task scores & \(\operatorname{mean}(T_1,\ldots,T_7)\) \\
LongMemEval & Recall@10, NDCG@10, Top-5 QA, Top-10 QA, Knowledge-Update QA & Mean of the five columns \\
RealPref & MCQ accuracy, T/F accuracy, preference awareness, preference alignment, answer quality & Mean of MCQ, T/F, and \(20\times\) each judge score \\
RealMem & Dynamic increase QA, dynamic update QA, proactive alignment QA, temporal reasoning QA & Mean of the four QA categories \\
CUPID & All instances F1, consistent F1, contrastive F1, changing F1, all instances generation score & Mean of the four F1 columns and \(10\times\) generation score \\
LoCoMo-Plus & LoCoMo-Plus score & Direct benchmark score \\
\bottomrule
\end{tabularx}
\end{table*}

\begin{figure*}[t]
\centering
\includegraphics[width=.88\textwidth]{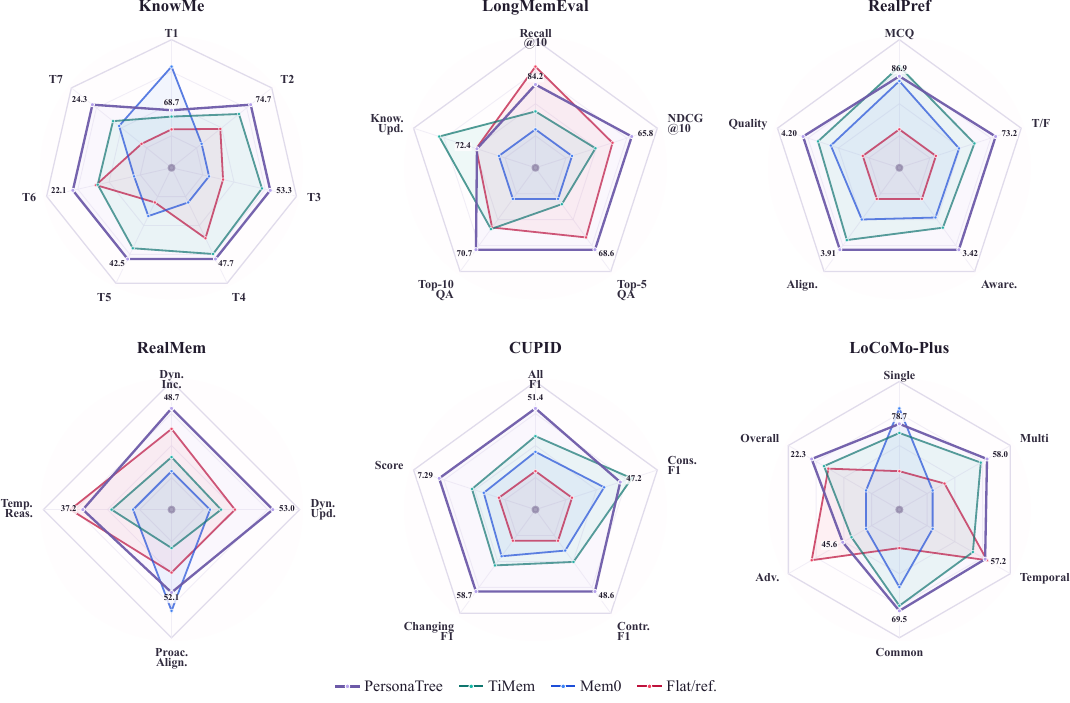}
\caption{\textbf{Benchmark metric breakdown under Qwen3-32B.}}
\label{fig:radar-breakdown}
\end{figure*}

Beyond the compact score, Figure~\ref{fig:radar-breakdown} reports the original benchmark metrics behind the Qwen3-32B results in Table~\ref{tab:main-results}. Each panel normalizes its own axes for readability, and the values next to the \method{} curve are the original metric values. The resulting profiles show how the aggregate gains are distributed across the metrics used by each benchmark.

The clearest gains for \method{} appear on CUPID and RealPref, where the curve extends outward on preference and generation axes such as alignment, answer quality, and changing preference F1. KnowMe shows a related result: the advantage is clearest on tasks T3 through T7, which require more abstraction over accumulated user evidence. RealMem and LoCoMo-Plus add positive evidence on dynamic update, multi hop, temporal, and overall axes, while LongMemEval shows strong QA and knowledge update results. Taken together, the six panels show that \method{} maintains a large overall advantage across person understanding, preference following, dynamic memory use, and long dialogue evaluation.

\subsection{Baseline and Inference Configuration}
\label{sec:appendix-baseline-config}

Comparisons are run through benchmark adapters that fix the split, answer backbone, output format, and judge protocol within each benchmark. Baseline specific memory construction remains inside each method's own interface; the controlled part is the downstream answering and scoring call.

\subsection{Benchmark Adapter Interfaces}
\label{sec:appendix-adapters}

Each benchmark adapter converts a benchmark instance into a common answer call without changing its native scoring protocol. The call contains the current query, task metadata, and the answer context supplied by the evaluated memory policy: full history for long context prompting, flat snippets for flat retrieval, or rendered support paths for \method{}. LongMemEval additionally scores retrieved items against answer-location labels with Recall@10 and NDCG@10. The adapter then normalizes outputs to the benchmark's required form, such as MCQ labels, true/false labels, short factual answers, open responses, or dialogue continuations, before invoking the original metric or judge.

\FloatBarrier
\input{full_results_tables}

\section{Qualitative Support-Path Examples}
\label{sec:appendix-qualitative-case}

The examples below show how support paths keep a benchmark query, the inferred preference, and the lower level evidence in the same retrieved context, complementing the ablation in Table~\ref{tab:full-path-ablation}.

\begin{casebox}{Example 1: Learning Environment Recommendation}
\casefield{User query} I'm interested in developing new skills---what kind of learning environment would you recommend?

\casefield{Flat retrieved nodes} A flat retriever may return several individually relevant memories: the user prefers live athlete interviews, dislikes guided tours, avoids distracting wearables, and enjoys social outdoor activities. Their relation to the learning query is left for the answer model to infer.

\casefield{\method{} support path}

\textit{Root.} The user values firsthand, social, and flexible experiences that preserve direct contact with people and activities.

\textit{Support-Mid.} The user prefers hands on, discussion based workshops or peer learning environments with live interaction and room for self direction.

\textit{Evidence-Leaf 1.} A live Q\&A with a retired football legend provided firsthand stories and practical wisdom that the user found especially insightful.

\textit{Evidence-Leaf 2.} An organized sightseeing tour felt too rigid; spontaneous visits to local sports venues felt more engaging.

\casefield{Path informed answer} The answer recommends a collaborative workshop with group projects and open discussion. The support path keeps the generalized learning preference and the observations that justify it together, so the recommendation follows the user's preference for firsthand, social, and flexible learning.
\end{casebox}

\begin{casebox}{Example 2: Work Environment Recommendation}
\casefield{User query} I'm looking for a new job---what kind of work environment should I consider?

\casefield{Flat retrieved nodes} A flat retriever may return memories about cycling, car commutes, animal shelter volunteering, and financial education. These nodes are relevant to the user, although the answer model must still infer which parts matter for the job recommendation.

\casefield{\method{} support path}

\textit{Root.} The user benefits from work arrangements with outdoor movement, flexible timing, and a low commute burden.

\textit{Support-Mid.} The user prefers flexible work schedules or remote options that allow midday outdoor breaks or exercise sessions.

\textit{Evidence-Leaf 1.} An hour long car commute left the user fatigued and frustrated, leading to an aversion to long car commutes.

\textit{Evidence-Leaf 2.} Endurance cycling became the user's main form of exercise because it combines physical challenge with peaceful time outdoors.

\casefield{Path informed answer} The answer recommends a remote or flexible role with autonomous scheduling and room for outdoor breaks. This recommendation reflects the user's need for outdoor movement and low commute burden.
\end{casebox}

\vspace{-0.35\baselineskip}
\section{Instruction Templates}
\label{sec:appendix-prompts}
\vspace{-0.55\baselineskip}

The templates below describe the model-facing instructions used for memory construction and retrieval. Each item lists the role, input, and structured output expected by the corresponding step. Benchmark answer formats follow the evaluation setup in Appendix~\ref{sec:appendix-evaluation}.

\begin{promptpanel}{Instructions Used by the Memory Pipeline}
\promptitem{Leaf extraction}{Role: expert cognitive memory extractor. Input: a recent conversation transcript. The model compresses the transcript into an objective event, the user's action or statement, and the user's feeling when supported by the text. Greetings and filler are removed, and the event summary is kept under 40 words. Objective fact or state updates receive a neutral emotion tag. Output fields are scene tags, emotion tag, and event summary.}

\promptitem{Mid pattern refinement}{Role: psychological behavior analyst. Input: a bounded list of related leaf events. The model summarizes one core behavioral pattern, stable state, or preference. Pattern types include condition trigger, trend change, frequency statistic, continuing state, and preference strength. The output contains pattern type, scene tags, emotion tendency, and pattern description.}

\promptitem{Root trait generation}{Role: psychological profiler. Input: established mid level patterns and optional case specific guardrails. The model produces one durable root claim in a calm third person declarative sentence. The claim is typed as personality, value, principle, objective identity, or hard boundary. Identity is reserved for objective roles or durable factual states, while hard boundary claims require explicit evidence such as an allergy, prohibition, intolerance, trauma trigger, or severe aversion.}

\promptitem{Root candidate grouping}{Role: organizer of mid level patterns into root candidates. Input: indexed mid nodes with pattern type, scene tags, emotion tendency, and pattern text. The model groups mids that reflect the same deeper trait, value, identity, principle, or boundary. Grouping follows latent coherence, and weakly related mids can remain ungrouped.}

\promptitem{Evidence validation}{Role: behavioral logic validator. Input: a known pattern and a new event. The model returns relevance and an ordered evidence category: strong support, weak support, weak conflict, or strong conflict. Unrelated pairs receive no evidence category. The category informs matching and confidence update.}

\promptitem{Query routing}{Role: retrieval router for a persona schema tree. Input: the user's latest query. The model predicts Leaf for specific events, details, or facts; Mid for habits, recent states, and preferences; and Root for hypothetical advice, core personality, or deeper user understanding. Output fields are target retrieval level and scene tags.}

\promptitem{Retrieved path rendering}{Input: a selected root, its supporting mids, and selected evidence leaves. The renderer exposes paths as \texttt{[Root]}, \texttt{[Support-Mid]}, and \texttt{[Evidence-Leaf]} blocks. These blocks enter the context for answer generation, keeping abstraction and supporting evidence adjacent.}
\end{promptpanel}

\begin{center}
\begin{minipage}{\columnwidth}
\centering
\setlength{\aboverulesep}{0pt}
\setlength{\belowrulesep}{0pt}
\setlength{\tabcolsep}{3.1pt}
\renewcommand{\arraystretch}{1.12}
\footnotesize
\captionof{table}{\textbf{Structured output schemas used by the memory pipeline.}}
\label{tab:prompt-output-schemas}
\begin{tabularx}{\columnwidth}{>{\bfseries\raggedright\arraybackslash}p{0.24\columnwidth}YY}
\toprule
\rowcolor{appendixhead}
\textbf{Component} & \textbf{Structured fields} & \textbf{Downstream use} \\
\midrule
Leaf extraction & Scene tags, emotion tag, event summary. & Creates event level evidence and supplies searchable descriptors for retrieval and consolidation. \\
Mid refinement & Pattern type, scene tags, emotion tendency, pattern description. & Forms reusable abstractions over compatible leaves and provides candidate parents for future evidence. \\
Root generation & Trait type and root description. & Promotes mature mid patterns into stable user claims used by higher level retrieval. \\
Root grouping & Candidate groups of mid node indices. & Selects semantically coherent mid clusters before root generation. \\
Evidence validation & Relevance flag and evidence category. & Updates match decisions and confidence through ordered support or conflict categories. \\
Query routing & Target retrieval level and scene tags. & Constrains retrieval to the abstraction depth and schema region requested by the query. \\
Benchmark answer & Task specific answer field. & Normalizes outputs for MCQ, true/false, short answer, open response, or continuation scoring. \\
\bottomrule
\end{tabularx}
\end{minipage}
\end{center}

\FloatBarrier
\section{Persona Schema Instantiation}
\label{sec:appendix-schema}

\begin{table*}[t]
\centering
\setlength{\aboverulesep}{0pt}
\setlength{\belowrulesep}{0pt}
\setlength{\tabcolsep}{5.0pt}
\renewcommand{\arraystretch}{1.2}
\small
\caption{\textbf{Persona schema fields in the experimental instantiation.}}
\label{tab:persona-schema-fields}
\begin{tabularx}{\textwidth}{>{\bfseries\raggedright\arraybackslash}p{0.10\textwidth}
                >{\raggedright\arraybackslash}p{0.25\textwidth}
                >{\raggedright\arraybackslash}p{0.30\textwidth}
                Y}
\toprule
\rowcolor{appendixhead}
\textbf{Level} & \textbf{Core fields} & \textbf{Schema descriptors} & \textbf{Main lifecycle role} \\
\midrule
Leaf & Event summary, timestamp, embedding, confidence, parent pointer. & Domain or scene tags, emotion tag, factual state, preference signal, boundary signal, role or schedule cue. & Stores local evidence from an interaction segment and supplies candidates for matching or consolidation. \\
Mid & Pattern description, pattern type, scene tags, emotion tendency, confidence, child leaves. & Condition trigger, trend change, frequency statistic, continuing state, preference strength. & Represents recurring states or patterns that connect multiple local events. \\
Root & Durable claim, trait type, confidence, supporting mids, support coverage. & Personality, value, principle, objective identity, or hard boundary. & Provides stable user claims for abstract person understanding and high level retrieval. \\
Edge & Parent-child relation, support direction, evidence category when available. & Support, conflict, or unrelated validation signal; temporal and schema compatibility. & Keeps higher level claims traceable to lower level evidence and carries validation outcomes into confidence update. \\
\bottomrule
\end{tabularx}
\end{table*}

The generic lifecycle uses a task schema to define node attributes and compatibility rules. In the person understanding setting, each leaf stores a compact memory text, temporal metadata, domain or scene descriptors, typed descriptors, an embedding, and a confidence score. The typed descriptors record roles, schedules, task facts, explicit preference statements, affective states, recurring triggers, values, and boundaries when these signals appear in the evidence. Before insertion, the normalization step rewrites factual updates as state descriptors and assigns neutral affect unless the interaction itself provides an emotional signal. This type assignment prevents durable factual states or constraints from being treated like ordinary preference evidence during decay, allowing persistent state descriptors to use reduced or zero decay while subjective preferences and affective patterns continue to fade without supporting evidence.

Mid and root schemas specify which abstractions can be formed from those leaves. The four broad mid types are preferences, ongoing states, recurring triggers, and temporal trends. During consolidation, orphan leaves are first partitioned by schema type and domain, then grouped by temporal proximity and embedding similarity. A group is accepted only when its member leaves share compatible descriptors and point to a consistent pattern. If the pattern matches an existing compatible mid, the member leaves attach to that mid and the description is rewritten from old and new support. Otherwise, the group is abstracted into a new mid node with its member leaves as support. Root nodes summarize compatible mids into identity, personality, values, principles, or safety relevant boundaries. A root promotion requires high confidence, multiple supporting leaves or mids, and coverage across time. Roots that represent constraints use conservative decay and can be retrieved as veto conditions when a query touches the corresponding boundary.

The evidence validator is applied after schema filtering. Its prompt contains the leaf text, the candidate mid description, and the relevant attributes, and returns support, conflict, or unrelated with a short rationale for auditing. Support and conflict outcomes guide the match score in Eq.~\ref{eq:match-score} and the confidence update in Eq.~\ref{eq:confidence-update}, while unrelated leaves remain unattached. At retrieval time, the predicted level and schema attributes filter candidates by domain, type, and boundary relevance. The renderer then builds alternative views of each support path, such as a root alone, a root with one supporting mid, or the full root--mid--leaf chain with selected leaves. The budgeted selection in Eq.~\ref{eq:budgeted-disclosure} decides which view enters the final context, so full paths are used only when the query needs that evidence depth and the token budget allows it.

\end{document}

%% file: full_results_tables.tex
\section{Ablation and Analysis Results}
\label{sec:appendix-full-analysis-results}

\paragraph{Hierarchy ablation.}
Table~\ref{tab:full-hierarchy-ablation} reports the task level KnowMe scores behind the hierarchy analysis. The comparison starts from retrieval over leaf evidence, then keeps refinement from leaves to mid nodes while root retrieval is disabled, and finally uses the complete PersonaTree hierarchy. The gains in macro average and in the more abstract KnowMe tasks show how root nodes complement local evidence for person understanding.

\begin{center}
\begin{minipage}{\columnwidth}
\centering
\setlength{\aboverulesep}{0pt}
\setlength{\belowrulesep}{0pt}
\renewcommand{\arraystretch}{1.18}
\captionof{table}{\textbf{Full KnowMe hierarchy ablation.}}
\label{tab:full-hierarchy-ablation}
\resizebox{\linewidth}{!}{
\begin{tabular}{>{\bfseries\raggedright\arraybackslash}lccccccccc}
\toprule
\rowcolor{appendixhead}
\textbf{Variant} & \textbf{T1} & \textbf{T2} & \textbf{T3} & \textbf{T4} & \textbf{T5} & \textbf{T6} & \textbf{T7} & \textbf{Avg} & \textbf{$\Delta$} \\
\midrule
Leaf Only & 66.9 & 73.4 & 45.9 & 44.8 & 36.7 & 19.5 & 16.2 & 43.3 & +0.0 \\
No Root & 67.4 & 74.0 & 51.7 & 47.5 & 41.9 & 20.1 & 18.8 & 45.9 & +2.6 \\
\method{} & \best{68.7} & \best{74.7} & \best{53.3} & \best{47.7} & \best{42.5} & \best{22.1} & \best{24.3} & \best{47.6} & \best{+4.3} \\
\bottomrule
\end{tabular}}
\end{minipage}
\end{center}

\paragraph{Path retrieval ablation.}
Table~\ref{tab:full-path-ablation} keeps the same Leaf, Mid, and Root inventory and changes only the retrieval policy. With memory content held fixed, the score and context differences reflect how evidence is organized in the retrieved context. MCQ and T/F are RealPref classification metrics; Aware., Align., and Qual. are judge scores for preference awareness, preference alignment, and answer quality; Comp. is the RealPref composite used in this paper, and Ctx. is the final answer context size.

\begin{center}
\begin{minipage}{\columnwidth}
\centering
\setlength{\aboverulesep}{0pt}
\setlength{\belowrulesep}{0pt}
\renewcommand{\arraystretch}{1.18}
\captionof{table}{\textbf{Full RealPref path retrieval ablation.}}
\label{tab:full-path-ablation}
\resizebox{\linewidth}{!}{
\begin{tabular}{>{\bfseries\raggedright\arraybackslash}lccccccc}
\toprule
\rowcolor{appendixhead}
\textbf{Variant} & \textbf{MCQ} & \textbf{T/F} & \textbf{Aware.} & \textbf{Align.} & \textbf{Qual.} & \textbf{Comp.} & \textbf{Ctx.} \\
\midrule
Flat Nodes & 86.5 & 72.4 & 3.38 & 3.61 & 4.00 & 75.7 & 3,180 \\
\method{} Path & \best{86.9} & \best{73.2} & \best{3.42} & \best{3.91} & \best{4.20} & \best{78.1} & \best{2,990} \\
\bottomrule
\end{tabular}}
\end{minipage}
\end{center}

\paragraph{Layer activation corpus.}
Table~\ref{tab:full-layer-activation} summarizes the activation analysis over 3 character histories, 27,062 timeline events, and 2,580 question records. The rows connect KnowMe task types with the memory levels activated by retrieval. Factual questions rely more on Leaf nodes, temporal and narrative questions shift toward Mid nodes, and insight questions activate Root nodes more often. Qs. is the number of questions, Dom. is the dominant activated level, Hits is the average number of retrieved nodes, and Ctx. is the average final context size.

\begin{center}
\begin{minipage}{\columnwidth}
\centering
\setlength{\aboverulesep}{0pt}
\setlength{\belowrulesep}{0pt}
\renewcommand{\arraystretch}{1.18}
\captionof{table}{\textbf{Full KnowMe layer activation statistics.} Leaf, Mid, and Root columns report activation percentages; rows sum to 100\%.}
\label{tab:full-layer-activation}
\resizebox{\linewidth}{!}{
\begin{tabular}{>{\bfseries\raggedright\arraybackslash}l>{\raggedright\arraybackslash}lccccccc}
\toprule
\rowcolor{appendixhead}
\textbf{Task} & \textbf{Level} & \textbf{Qs.} & \textbf{Leaf} & \textbf{Mid} & \textbf{Root} & \textbf{Dom.} & \textbf{Hits} & \textbf{Ctx.} \\
\midrule
T1 & Level I / Fact \& Entity & 330 & 73.9 & 18.2 & 7.9 & Leaf & 6.72 & 2,694 \\
T2 & Level I / Fact \& Entity & 300 & 58.0 & 34.0 & 8.0 & Leaf/Mid & 7.31 & 2,604 \\
T3 & Level I / Fact \& Entity & 561 & 43.0 & 45.1 & 11.9 & Leaf/Mid & 7.55 & 3,187 \\
T4 & Level II / Temporal Logic & 125 & 28.8 & 66.4 & 4.8 & Mid & 7.09 & 2,993 \\
T5 & Level II / Temporal Logic & 581 & 36.0 & 58.0 & 6.0 & Mid & 7.52 & 3,341 \\
T6 & Level III / Insight & 495 & 22.0 & 30.9 & 47.1 & Root & 8.40 & 3,744 \\
T7 & Level III / Insight & 188 & 18.1 & 38.8 & 43.1 & Root & 7.87 & 3,816 \\
\bottomrule
\end{tabular}}
\end{minipage}
\end{center}